\documentclass[journal]{IEEEtran}
\usepackage{amsmath}
\usepackage{graphicx}
\usepackage{pifont}
\usepackage{multirow}
\usepackage{cite}
\newcommand{\cmark}{\ding{51}}
\newcommand{\xmark}{\ding{55}}
\ifCLASSINFOpdf

\else

\fi

\begin{document}

\title{FedMAD: Modulation-Aware Directional Aggregation for Federated Learning in Remote Sensing Image Classification}

\author{Barış~Büyüktaş,~\IEEEmembership{Member,~IEEE,}
        and~Begüm~Demir,~\IEEEmembership{Senior Member,~IEEE}
        
\thanks{B. Büyüktaş and B. Demir are with the Faculty of Electrical Engineering and Computer Science, Technische Universität Berlin, 10587 Berlin, Germany, also with the BIFOLD - Berlin Institute for the Foundations of Learning and Data, 10587 Berlin, Germany. Email: baris.bueyuektas@tu-berlin.de, demir@tu-berlin.de.}
}

\maketitle

\markboth{Journal of \LaTeX\ Class Files,~Vol.~13, No.~9, September~2014}%
{Shell \MakeLowercase{\textit{et al.}}: Bare Demo of IEEEtran.cls for Journals}

\maketitle

\begin{abstract}
Federated learning (FL) has recently attracted increasing attention in remote sensing (RS) since it enables collaborative model training across decentralized RS image archives without requiring direct access to local data. However, FL performance significantly degrades when the data distributions between clients are heterogeneous, which often occurs due to geographical differences, seasonal changes, and varying image acquisition and atmospheric conditions. To address this challenge, in this letter, we propose a novel personalized FL framework (denoted as FedMAD) for RS image classification problems. The proposed framework separates globally shared representation parameters from client-specific adaptation parameters to preserve client-specific features while maintaining globally transferable representations. This is achieved by integrating lightweight modulation modules and local batch normalization layers into the backbone network. Although globally shared parameters are collaboratively optimized between clients, client-specific parameters remain local to preserve domain-specific feature characteristics. In addition, FedMAD introduces a modulation-aware directional aggregation strategy that dynamically adjusts the importance of aggregation for each client according to the alignment of local modulation updates. This allows the global optimization process to suppress conflicting client updates originating from heterogeneous data distributions while enhancing the contribution of clients with consistent adaptation behaviors. The experimental results obtained on the BigEarthNet-S2 and EuroSAT datasets demonstrate the effectiveness of FedMAD compared to state-of-the-art FL algorithms under heterogeneous RS data distributions. The code of the proposed framework will be publicly available at https://git.tu-berlin.de/rsim/fedmad.
\end{abstract}


\IEEEpeerreviewmaketitle

\section{Introduction}

\IEEEPARstart{F}{ederated} learning (FL) has emerged as a promising learning paradigm for collaborative training of deep neural networks across decentralized remote sensing (RS) image archives (i.e., clients) without requiring direct access to local training data \cite{tan2025towards,buyuktacs2026fedx}. In FL, each client locally optimizes its model using local data and only exchanges model parameters with a central server during training. Therefore, FL can be used for many RS applications where transferring all local data to a centralized server may be infeasible due to privacy concerns, legal regulations, communication limitations, and storage constraints \cite{buyuktacs2024federated}.

Although FL offers significant advantages for decentralized RS applications, its performance substantially degrades when data distributions between clients are heterogeneous (i.e., non-IID data) \cite{li2025distributed}. Non-IID data are particularly common in RS, as image characteristics can vary significantly due to differences in image acquisition conditions \cite{tuia2024artificial}. Under heterogeneous data distributions, conventional FL algorithms that aggregate models solely according to the number of local training samples often suffer from unstable convergence and reduced generalization capability \cite{li2021fedbn}.

To address the challenges caused by heterogeneous data distributions, several personalized FL algorithms have recently been introduced in computer vision and machine learning communities \cite{zheng2025fedcalm,yi2026pfedmoe}. Existing approaches commonly aim to preserve client-specific information through local normalization layers, personalized model branches, or client-specific classifier heads \cite{sabah2024model}. Although these methods improve the capability of FL models to adapt to client-specific data distributions, most existing studies employ conventional aggregation strategies that equally treat client updates during global optimization. Under highly heterogeneous RS distributions, different clients may contribute inconsistently to the global optimization process. As a result, aggregating conflicting local updates may negatively affect the convergence and generalization capability of the global model \cite{lu2024federated}.

To alleviate the limitations of existing personalized FL algorithms, we propose a novel FL framework (denoted as FedMAD) for RS image classification. FedMAD integrates client-specific modulation learning with an adaptive aggregation strategy to simultaneously preserve local feature characteristics and account for differences in client adaptation behaviors during global optimization. To this end, lightweight modulation modules and local batch normalization layers are utilized to preserve domain-specific feature characteristics while collaboratively optimizing globally transferable representations across clients. Moreover, the proposed framework dynamically estimates the importance of client aggregation according to the consistency of local modulation updates, allowing the global optimization process to suppress conflicting local updates. Experimental analyses conducted on the BigEarthNet-S2 \cite{clasen2024reben} and EuroSAT \cite{helber2019eurosat} datasets demonstrate the effectiveness of the proposed framework compared with state-of-the-art FL algorithms under heterogeneous RS data distributions. The main contributions of this letter are summarized as follows:
\begin{itemize}
\item We propose a personalized FL framework for RS image classification that separates globally shared parameters from client-specific adaptation parameters.
\item We introduce lightweight modulation modules integrated with local batch normalization layers to preserve domain-specific feature characteristics across heterogeneous RS image archives.
\item We propose a modulation-aware directional aggregation strategy that dynamically adjusts the importance of client aggregation according to the alignment of local adaptation updates.

\end{itemize}

\section{Proposed FedMAD Framework}

The proposed FedMAD framework aims to address the limitations caused by heterogeneous RS data distributions during FL training. To this end, FedMAD consists of two main components: 1) client-specific representation learning and 2) modulation-aware directional aggregation. The details of these components are presented in the following subsections.

\subsection{Client-Specific Representation Learning}

\begin{figure}[t]
\centering
\centerline{\includegraphics[width=1\columnwidth,keepaspectratio]{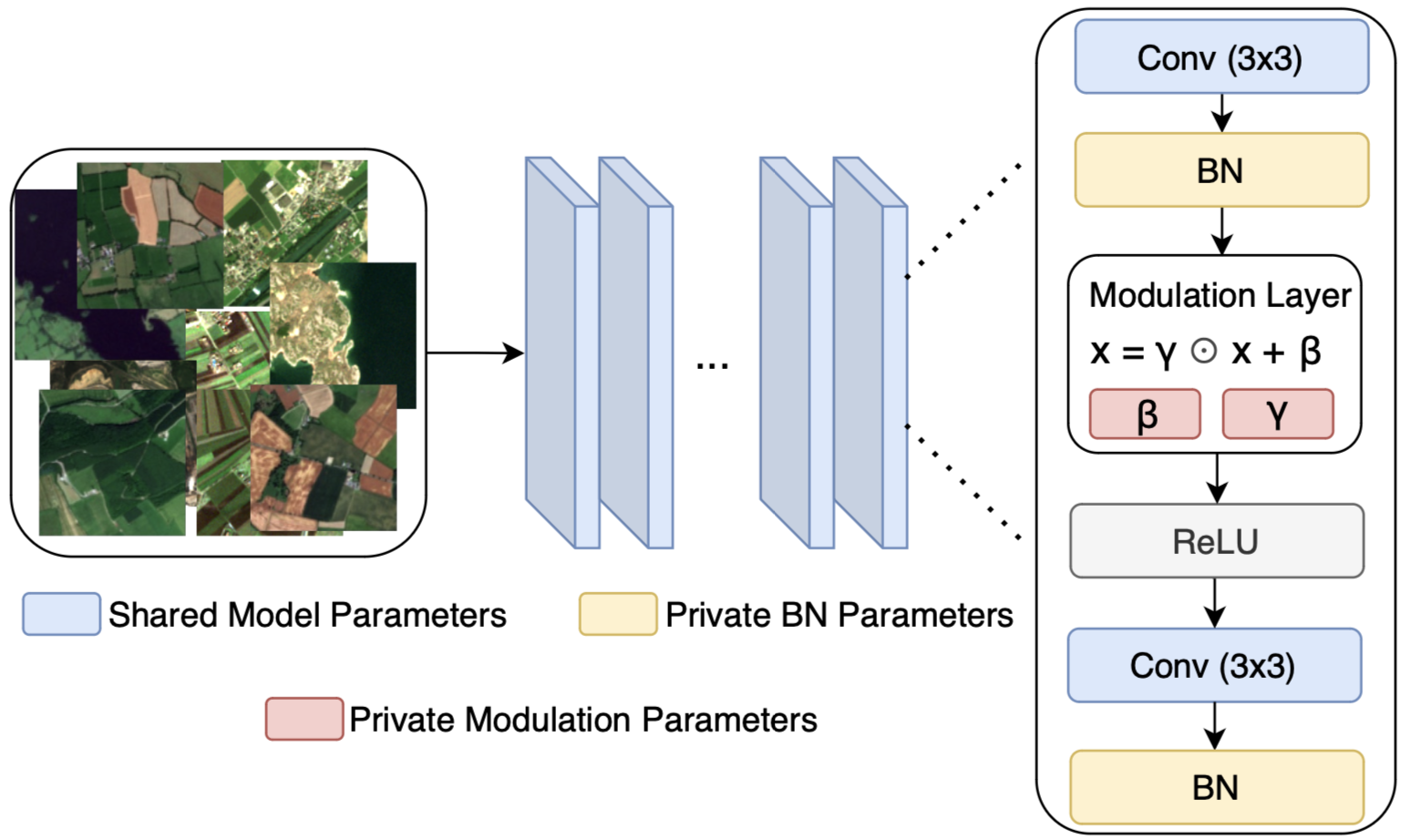}}
\caption{Illustration of the globally shared and client-specific model parameters in FedMAD.}
\label{figure}
\end{figure}

Let $K$ denote the total number of clients and $\mathcal{C}_i$ denote the $i$th client for $1 \leq i \leq K$. Each client $\mathcal{C}_i$ is associated with a local training set $\mathcal{D}_i = \{(\mathbf{x}_i^z,\mathbf{y}_i^z)\}_{z=1}^{M_i}$, where $M_i$ is the number of training samples, $\mathbf{x}_i^z$ denotes the $z$th RS image, and $\mathbf{y}_i^z$ represents the corresponding image annotation. To preserve domain-specific feature characteristics across heterogeneous clients, FedMAD separates globally shared representation parameters from client-specific adaptation parameters during FL training. Let $\mathbf{w}_g$ denote the globally shared parameters and $\mathbf{w}_i^p$ denote the client-specific parameters associated with the client $\mathcal{C}_i$. Consequently, the parameters of the local model can be represented as follows:
\begin{equation}
\mathbf{w}_i = \{\mathbf{w}_g,\mathbf{w}_i^p\}.
\end{equation}
This design enables the model to learn transferable image representations across clients while preserving local adaptations that are specific to individual data distributions. Fig.~\ref{figure} illustrates the separation of globally shared and client-specific parameters in the proposed FedMAD framework. To this end, lightweight modulation modules are integrated into the backbone network to enable adaptation of client-specific characteristics while maintaining globally transferable image representations. Each modulation module performs the feature-wise affine transformation as follows:
\begin{equation}
\mathbf{y} = \boldsymbol{\gamma} \odot \mathbf{x} + \boldsymbol{\beta},
\end{equation}
where $\mathbf{x}$ and $\mathbf{y}$ denote the input and output feature maps, respectively, $\odot$ represents element-wise multiplication, and $\boldsymbol{\gamma}$ and $\boldsymbol{\beta}$ denote learnable modulation parameters. The modulation parameters dynamically adjust feature activations according to the local data distribution of each client, allowing personalized feature adaptation under heterogeneous RS image distributions. As a result, different clients can adapt the shared feature representations to their own domain characteristics without modifying the overall backbone architecture.

To stabilize local feature statistics during FL training, the modulation modules are integrated before the batch normalization (BN) layers of the backbone network. In addition, the BN parameters and modulation parameters are preserved locally and excluded from global aggregation. Accordingly, the client-specific parameters $\mathbf{w}_i^p$ consist of the modulation and BN parameters associated with the client $\mathcal{C}_i$, while the remaining backbone parameters are collaboratively optimized across clients as globally shared parameters $\mathbf{w}_g$. This design allows FedMAD to preserve regional image characteristics and domain-specific feature statistics without disrupting the learning of globally transferable representations. At the same time, the number of client-specific parameters remains relatively small compared to the overall size of the network. Consequently, personalized feature adaptation can be achieved without introducing a noticeable computational overhead.

For each communication round, every client locally updates its model parameters using its corresponding local training set. Let $\phi_i(\cdot;\mathbf{w}_i)$ denote the local model associated with the client $\mathcal{C}_i$. The local objective function is defined as follows:
\begin{equation}
\mathcal{L}_i(\mathbf{w}_i)=
\sum_{(\mathbf{x}_i^z,\mathbf{y}_i^z)\in\mathcal{D}_i}
\ell(\phi_i(\mathbf{x}_i^z;\mathbf{w}_i),\mathbf{y}_i^z),
\end{equation}
where $\ell(\cdot)$ denotes the classification loss function. During local optimization, both the globally shared and client-specific parameters are updated using local training data. After local training, only the globally shared parameters are transmitted to the central server for aggregation, whereas the client-specific parameters remain local throughout FL training. This strategy allows each client to retain personalized feature representations while simultaneously benefiting from the knowledge acquired across all participating clients.

\subsection{Modulation-Aware Directional Aggregation}

Most FL algorithms aggregate local model updates according to the number of training samples associated with each client. However, in heterogeneous RS data distributions, local optimization directions may differ significantly between clients. This may result in conflicting updates during global optimization. To alleviate this limitation, FedMAD dynamically adjusts the importance of aggregation of each client according to the alignment of local modulation updates.

Let $\mathbf{m}_i^t$ and $\mathbf{m}_i^{t+1}$ denote the modulation parameters of the client $\mathcal{C}_i$ before and after local optimization at the communication round $t$, respectively. The local modulation update is computed as follows:
\begin{equation}
\Delta \mathbf{m}_i =
\mathbf{m}_i^{t+1} - \mathbf{m}_i^t.
\end{equation}
Then, the normalized modulation update direction associated with the client $\mathcal{C}_i$ is obtained as follows:
\begin{equation}
\mathbf{u}_i =
\frac{\Delta \mathbf{m}_i}
{\|\Delta \mathbf{m}_i\|_2}.
\end{equation}
To estimate the dominant global adaptation behavior across clients, a reference modulation direction is computed by aggregating all local modulation directions as follows:
\begin{equation}
\mathbf{u}_{ref} =
\frac{\sum_{i=1}^{K}\mathbf{u}_i}
{\left\|\sum_{i=1}^{K}\mathbf{u}_i\right\|_2}.
\end{equation}
Then, the alignment score associated with the client $\mathcal{C}_i$ is calculated according to the cosine similarity between the local modulation direction and the reference direction as follows:
\begin{equation}
a_i =
\frac{\mathbf{u}_i^\top \mathbf{u}_{ref}}
{\|\mathbf{u}_i\|_2\|\mathbf{u}_{ref}\|_2}.
\end{equation}
The proposed aggregation strategy utilizes modulation updates, since modulation parameters directly characterize client-specific adaptation behaviors under heterogeneous RS data distributions. Since these parameters are exclusively responsible for the adaptation of local features, their update directions provide valuable information on how each client adjusts its representations to its own data distribution. Therefore, the alignment between local modulation updates provides an effective indicator of the consistency between client optimization behaviors during FL training. Based on alignment scores, FedMAD dynamically estimates the importance of aggregation of each client during global optimization. To this end, the aggregation weight associated with the client $\mathcal{C}_i$ is computed as follows:
\begin{equation}
\alpha_i =
\frac{
\frac{|\mathcal{D}_i|}{|\mathcal{D}|}
(a_i+\epsilon)^p
}
{
\sum_{j=1}^{K}
\frac{|\mathcal{D}_j|}{|\mathcal{D}|}
(a_j+\epsilon)^p
},
\end{equation}
where $|\mathcal{D}_i|$ denotes the number of training samples associated with the client $\mathcal{C}_i$, $\epsilon$ is a small constant for numerical stability, and $p$ controls the sensitivity of the aggregation process to the alignment scores. Finally, the globally shared parameters are updated on the central server as follows:
\begin{equation}
\mathbf{w}_g^{t+1}
=
\sum_{i=1}^{K}
\alpha_i
\mathbf{w}_{g,i}^{t+1},
\end{equation}
where $\mathbf{w}_{g,i}^{t+1}$ denotes the locally optimized globally shared parameters associated with the client $\mathcal{C}_i$. By dynamically suppressing conflicting local updates while emphasizing clients with consistent adaptation behaviors, the proposed aggregation strategy is capable of enhancing the robustness of FL optimization under non-IID data.

\section{Experimental Results}

\subsection{Dataset Description}

In the experiments, we selected the following datasets to evaluate the proposed FedMAD framework: 1) BigEarthNet-S2 v2.0 \cite{clasen2024reben}; and 2) EuroSAT \cite{helber2019eurosat}. BigEarthNet-S2 is a large-scale multilabel RS image dataset consisting of Sentinel-2 image patches associated with multiple land-cover classes. In the experiments, we utilized the 10m and 20m bands, whereas the 60m bands were not considered. The 20m bands were resized to 120×120 pixels using bicubic interpolation, resulting in ten bands for each image. To simulate heterogeneous FL setup, clients were constructed according to countries, where each client only contained images acquired from a specific country. In the experiments, we considered Finland, Ireland, Serbia, Austria, Belgium, Lithuania, Portugal, and Switzerland as individual clients. In addition, only summer-season images were utilized for reducing seasonal inconsistencies across clients.

The EuroSAT dataset consists of Sentinel-2 image patches associated with 10 scene classes representing different land-cover and land-use categories. In the experiments, only the RGB bands were utilized. Each image has a spatial size of $64 \times 64$ pixels. We randomly divided the dataset into training and test sets with a ratio of 80\% and 20\%, respectively. To simulate heterogeneous FL environments, the training samples were distributed across clients using a Dirichlet distribution with concentration parameter $\alpha = 0.3$. This results in non-IID label distributions among clients. The same test set was utilized for evaluating all FL algorithms throughout the experiments.

\subsection{Design of Experiments}

In the experiments, we considered ResNet50 as the backbone architecture. We analyzed the sensitivity of the proposed framework with respect to the number of modulation layers and the aggregation parameter $p$. For this purpose, the number of modulation layers was varied from 1 to 16, while the aggregation parameter $p$ was varied from 0.5 to 3.0. The corresponding results are presented in Section \ref{sens}. Based on the obtained results, the best-performing configuration was selected and subsequently used in the comparison with the state-of-the-art FL algorithms. For all experiments, the number of communication rounds was set to 40, while each client performed 2 local training epochs per communication round to provide sufficient local optimization while maintaining efficient FL training. All models were trained using the Adam optimizer with a learning rate of 0.001 and a batch size of 512. The performance was evaluated using micro mean Average Precision (mAP) for BigEarthNet-S2 and overall accuracy (OA) for EuroSAT. In addition, the computational complexity of the considered methods was assessed in terms of floating-point operations (FLOPs).

\subsection{Ablation and Sensitivity Study}
\label{sens}

In this subsection, we analyze the contribution of the individual components of the proposed FedMAD framework and assess its sensitivity to key hyperparameters. Tables~\ref{tab:ablation} and~\ref{tab:sensitivity} show the corresponding results.
From the ablation results, one can observe that each component contributes positively to the overall performance. Starting from the FedBN baseline, which utilizes only private batch normalization layers, the introduction of the proposed modulation-aware aggregation strategy improves micro mAP from 72.42\% to 73.43\% on BigEarthNet-S2 and from 56.50\% to 59.82\% on EuroSAT. These improvements indicate that dynamically weighting client contributions according to the alignment of local adaptation behaviors leads to a more effective global optimization process under heterogeneous data distributions. Introducing private modulation modules while employing conventional aggregation further improves the performance to 73.91\% on BigEarthNet-S2 and 64.67\% on EuroSAT. This behavior demonstrates the effectiveness of the proposed client-specific representation learning strategy in preserving domain-specific feature characteristics across heterogeneous clients. The best performance is achieved when all components are jointly utilized. In particular, the complete FedMAD framework achieves 75.12\% micro mAP on BigEarthNet-S2 and 67.14\% on EuroSAT, demonstrating that the proposed client-specific adaptation mechanism and modulation-aware aggregation strategy provide complementary benefits.

Table~\ref{tab:sensitivity} further analyzes the influence of the number of modulation layers and the alignment power parameter $p$ on the BigEarthNet-S2 dataset. From the results, one can observe that increasing the number of modulation layers initially improves the performance, reaching the best micro mAP score of 75.12\% when four modulation layers are employed. This behavior suggests that incorporating client-specific adaptation into multiple stages of the network enables more effective modeling of heterogeneous RS data distributions. However, a larger number of modulation layers leads to a performance decrease, suggesting that excessive client-specific adaptation may reduce the effectiveness of global knowledge sharing between clients. Regarding the aggregation parameter $p$, the best performance is obtained when $p=2.0$. Smaller values result in weaker differentiation between client contributions, whereas larger values overly favor highly aligned clients and diminish the contribution of other clients. Furthermore, the relatively small performance variations observed for the $p$ values between 0.5 and 2.0 indicate that the proposed aggregation strategy remains stable under moderate changes in the alignment sensitivity parameter. Overall, the results demonstrate that the proposed FedMAD framework is robust to moderate variations in its hyperparameters while consistently benefiting from both client-specific adaptation and modulation-aware aggregation.

\begin{table}[t!]
\renewcommand{\arraystretch}{1.4}
\caption{Ablation study of the proposed FedMAD framework. The evaluated components include private batch normalization (BN) layers, private modulation (Mod.) layers, and the modulation-aware (MA) aggregation strategy. Performance is reported on the BigEarthNet-S2 and EuroSAT datasets.}
\centering
\label{tab:ablation}
\begin{tabular}{@{}ccc@{\hspace{20pt}}cc@{}}
\hline
\multicolumn{3}{@{\hspace{-10pt}}c}{\textbf{Modules}} &
\multicolumn{2}{c}{\textbf{Performance (\%)}} \\
\textbf{Private} & \textbf{Private} & \textbf{MA} &
\textbf{BigEarthNet-S2} & \textbf{EuroSAT} \\
\textbf{BN} & \textbf{Mod.} & \textbf{Aggregation} & & \\
\hline
\cmark & \xmark & \xmark & 72.42 & 56.50 \\
\cmark & \xmark & \cmark & 73.43 & 59.82 \\
\cmark & \cmark & \xmark & 73.91 & 64.67 \\
\cmark & \cmark & \cmark & \textbf{75.12} & \textbf{67.14} \\
\hline
\end{tabular}
\end{table}

\begin{table}[ht]
\renewcommand{\arraystretch}{1.3}
\caption{Sensitivity analysis of the proposed FedMAD framework in terms of the number of modulation layers and the alignment power parameter $p$ on the BigEarthNet-S2 dataset.}
\centering
\label{tab:sensitivity}
\begin{tabular}{lcc}
\hline
\textbf{Setting} & \textbf{Configuration} & \textbf{Micro mAP (\%)} \\
\hline
\multirow{5}{*}{Modulation layers}
& 1 & 70.24 \\
& 2 & 73.95 \\
& 4 & \textbf{75.12} \\
& 8 & 71.28 \\
& 16 & 71.49 \\
\hline
\multirow{4}{*}{$p$}
& 0.5 & 73.52 \\
& 1.0 & 74.39 \\
& 2.0 & \textbf{75.12}  \\
& 3.0 & 71.43 \\
\hline
\end{tabular}
\end{table}

\subsection{Comparison with State-of-the-Art FL Methods}

In this subsection, we compare the proposed FedMAD framework with FedAvg and five state-of-the-art FL algorithms: MOON, FedDC, FedAWA, FedGPS, and FedBN. Table~\ref{tab:sota} shows the corresponding accuracies obtained by the considered algorithms.
From the results, one can observe that the proposed FedMAD framework consistently achieves the highest performance on both datasets. In particular, FedMAD achieves a micro mAP score of 75.12\% on BigEarthNet-S2 and an overall accuracy of 67.14\% on EuroSAT. In particular, the proposed framework outperforms FedAvg by approximately 14\% on both datasets. These results indicate that preserving client-specific feature characteristics and dynamically suppressing conflicting local updates substantially improve FL optimization under heterogeneous RS data distributions. In greater detail, MOON and FedDC improve the performance compared with FedAvg by introducing additional regularization mechanisms to stabilize local optimization under heterogeneous client distributions. Similarly, FedAWA and FedGPS achieve stronger performance by dynamically adjusting the aggregation process according to client behaviors during FL training. However, although these algorithms improve the robustness of FL optimization, they still globally aggregate the entire model representation without explicitly separating globally transferable representations from client-specific feature adaptations. In contrast, the proposed FedMAD framework preserves local modulation and batch normalization parameters on each client while collaboratively optimizing globally shared representations. This enables the framework to better preserve regional feature characteristics and domain-specific feature statistics across heterogeneous RS clients. FedBN achieves competitive performance on BigEarthNet-S2 due to its capability to preserve local feature statistics using client-specific batch normalization layers. However, the proposed FedMAD framework further improves the performance by integrating lightweight modulation modules together with the proposed modulation-aware directional aggregation strategy. In particular, the improvement over FedBN demonstrates that preserving only local normalization statistics is insufficient for effectively modeling heterogeneous RS distributions. By utilizing modulation-update alignment during global aggregation, FedMAD can dynamically suppress conflicting client updates while emphasizing clients with consistent adaptation behaviors. This results in improved convergence and stronger generalization capability under heterogeneous FL settings. Overall, the obtained results demonstrate that the proposed framework consistently improves the robustness and generalization capability of FL optimization compared with the considered state-of-the-art FL algorithms for RS image classification.

In addition to the classification performance, Table~\ref{tab:sota} shows that the computational complexity of FedMAD remains comparable to that of the considered FL algorithms. Although FedMAD introduces additional modulation modules, the resulting increase in FLOPs is negligible due to the lightweight nature of the feature-wise affine transformations. As a result, the computational complexity of FedMAD remains very close to that of FedAvg, FedBN, FedDC, and FedAWA. In contrast, MOON and FedGPS exhibit slightly higher FLOPs owing to the additional projection and prototype heads employed during training. These results demonstrate that the proposed FedMAD framework achieves substantial performance gains while introducing only a negligible computational overhead.

\begin{table}[!t]
\renewcommand{\arraystretch}{1.4}
\caption{Micro mAP (\%), OA (\%), and FLOPs (G) obtained by the proposed FedMAD framework and the considered state-of-the-art FL algorithms on the BigEarthNet-S2 and EuroSAT datasets.}
\centering
\label{tab:sota}
\begin{tabular}{lccc}
\hline
\textbf{Algorithm} & \multicolumn{2}{c}{\textbf{Performance (\%)}} & \textbf{FLOPs (G)} \\
\cline{2-3}
& \textbf{BigEarthNet-S2} & \textbf{EuroSAT} & \\
\hline
FedAvg & 60.78 & 54.12 & 2.7021 \\
MOON~\cite{li2021model} & 65.93 & 63.77 & 2.7033 \\
FedDC~\cite{gao2022feddc} & 66.43 & 64.03 & 2.7021 \\
FedAWA~\cite{shi2025fedawa} & 67.33 & 64.44 & 2.7021 \\
FedGPS~\cite{yang2026fedgps} & 70.86 & 64.95 & 2.7032 \\
FedBN~\cite{li2021fedbn} & 72.42 & 56.50 & 2.7021 \\
Ours & \textbf{75.12} & \textbf{67.14} & 2.7029 \\
\hline
\end{tabular}
\end{table}

\section{Conclusion}

In this letter, we have proposed FedMAD, a novel personalized FL framework for RS image classification under heterogeneous data distributions. The proposed framework separates globally shared representation parameters from client-specific adaptation parameters through lightweight modulation modules and local batch normalization layers. In addition, a modulation-aware directional aggregation strategy is introduced to dynamically adjust client aggregation importance according to the alignment of local adaptation behaviors. Experimental results obtained on the BigEarthNet-S2 and EuroSAT datasets demonstrate that the proposed framework consistently outperforms state-of-the-art FL algorithms while introducing only a negligible computational overhead. Since the proposed modulation modules and aggregation strategy are independent of the network architecture, FedMAD can be integrated into different DL backbones and FL systems. Future work will explore adaptive mechanisms for dynamically determining the degree of personalization and knowledge sharing during FL.


\ifCLASSOPTIONcaptionsoff
  \newpage
\fi

\bibliographystyle{IEEEtran}
{\small
\bibliography{refs}}

\end{document}